# Cloud-based user modeling for social robots: a first attempt

Marco Botta
Computer Science Dept.
University of Turin, Italy
marco.botta@unito.it

Daniele Camilleri
MSc in Computer Science
University of Turin, Italy
daniele.camilleri@edu.unito.it

Federica Cena
Computer Science Dept.
University of Turin, Italy
federica.cena@unito.it

Francesco Di Sario
MSc in Computer Science
University of Turin, Italy
francesco.disario@edu.unito.it

Cristina Gena
Computer Science Dept.
University of Turin, Italy
cristina.gena@unito.it

Giuseppe Ignone
MSc in Computer Science
University of Turin, Italy
giuseppe.ignone@edu.unito.it

Claudio Mattutino
Computer Science Dept.
University of Turin, Italy
claudio.mattutino@unito.it

**Abstract**

A social robot is an autonomous robot that interact with people by engaging in socialemotive behaviors, skills, capacities, and rules attached to its collaborative role. In order to achieve these goals we believe that modeling the interaction with the user and adapt the robot behavior to the user herself are fundamental for its social role. This paper presents our first attempt to integrate user modeling features in social and affective robots. We propose a cloud-based architecture for modeling the user-robot interaction in order to reuse the approach with different kind of social robots.

Human Robot Interaction, User modeling, Affective Interaction

## 1 Introduction

A social robot is an autonomous robot that interact with people by engaging in social-emotive behaviors, skills, capacities, and rules related to its collaborative role [1]. According to Fong et al. [5], social robots should:

- express and/or perceive emotions;
- communicate with high-level dialogue;
- use natural cues (gaze, gestures, etc.);
- exhibit distinctive personality and character;
- learn/recognize models of other agents;
- establish/maintain social relationships;
- learn/develop social competencies;
- communicate both in verbal and not verbal way.

Thus, social interactive robots need to perceive and understand the richness and complexity of user's natural social behavior, in order to interact with people in a human-like manner [1]. Detecting and recognizing human action and communication provides a good starting point, but more important is the ability to interpret and react to human behavior, and a key mechanism to carry out these actions is user modeling. User models are used for a variety of purposes by robots [5]. First, user models can help the robot understand an individual's behavior and dialogue. Secondly, user models are useful for adapting the behavior of the robot to the different abilities, experiences, knowledge, and preferential choices [19] of user. Finally, they determine the control form and feedback given to the user (for example, stimulating the interaction).

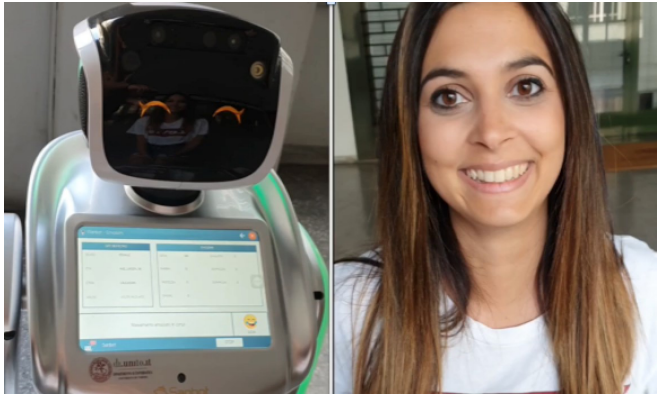

Figure 1: The iSanbot application for emotion detection.

This paper presents the our attempt to integrate user modeling features in social and affective robots. We propose a cloud-based architecture for modeling the user-robot interaction in order to re-use the approach with different kind of social robots.

## 2 Related work: User Modeling in Robotics

The field of user modeling in robotics is still in its infancy. Some preliminary works perform an initial questionnaire to categorize their users in stereotypes. [6] used stereotype to configure human-robot interaction, to adapt the dialogue and to modify the robot behavior. [2] employs a number of questionnaires, and then clusters the data to produce a number of stereotypes in form of personas. [4], which presents a companion robot that is adapted to the user by performing an initial questionnaire for configuration, More recently, [11, 12] exploits unsupervised methods to obtain the user's profiles.

Other work rely on implicit data gathering, such as [15], [7], [10]. Crucial is the ability of the robot to understand and model the user's emotion [16], [8].

The final goal of user modeling is the adaptation of the robot to the user features [13]. For example, the robot in [15] changes his task plan according to the user's plan, showing efficiency in task execution. [9] presents a robot tutoring system, which adapt himself to the user's preferences and skill level. [18] presents a robot guide that learns the user's preferences regarding its actions. Finally Mazzei et al. [14] proposed a personalized human-(virtual) robot experience for helping
people to have healthy life.

## 3 The iRobot application

The aim of our project is to develop a general purpose cloud-based application, called iRobot, offering cloud components for managing social, affective and adaptive services for social robots. This first version of the application has been developed and tested for the Sanbot Elf (henceforth simply Sanbot) robot acting as client, thus this client-side application has been called iSanbot. Thanks to the cloud-based service, the robot is able to handle a basic conversation with users, to recognized them and follow a set of social relationship rules, to detect the user's emotions and modify its behavior according to them. This last task is focused on real-time emotion detection and on how these could change human robot interaction. As case study, we started from a scenario wherein the Sanbot robot welcomes people who enter the hall of the Computer Science Dept. of the University of Turin. The robot must be able to recognize the people it has already interacted with and remember past interactions (mainly with the employees of the Dept.), introduce itself and be able to welcome new people who arrive and give them information about the offices. In the future, it must also be able to accompany people to offices and back to the entrance. As far as the offices information are concerned, we have developed a Sanbot application able to deliver this information to the user during a basic welcome conversation. This application will be no more detailed here, since this is out of the scope of the paper.

### 3.1 Sanbot description

Sanbot Elf[1] is a humanoid-shaped robot with a total size of 92 x 30 x 40 cm and a weight of 19 kg. The body structure is mostly made of plastic with a motor based on wheels that allows the robot to rotate over 360 degrees. Sanbot has a total of three cameras, two in the head and one in the chest belonging to the tablet incorporated in the robot. The two cameras in the head are the HD Camera and the 3D Camera, 8.0MP and 1.0MP respectively. The HD Camera can be used to perform tasks like taking pictures, recording videos and audio and camera's live streaming. In the Sanbot's head there is also a microphone, over three sensors located on the head and a projector located behind the head

---

[1] http://en.sanbot.com/product/sanbot-elf/

with a resolution of 1920x720 60hz 16:9. The body of Sanbot is equipped with a tablet incorporated on the chest, which can be used to show and use all the applications installed on it. The operating system is a modified version of Android, ergo the app development takes place on Android Studio and through the Sanbot SDK. Sanbot is equipped with sensors throughout all

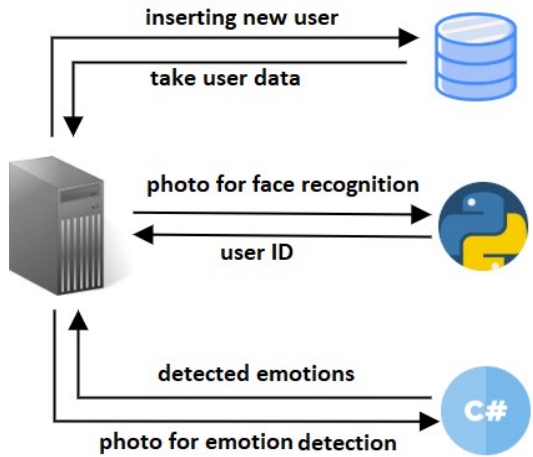

Figure 2: The iSanbot architecture. its body like infrared sensors, touch sensors and PIR, WIFI, Bluetooth and ZigBee wireless connections.

The robot is controllable via 7 modules, also called managers, which can be invoked with the use of API libraries. The 7 modules are:

1. Voice module.
2. Hardware module.
3. Hardware control module.
4. System Management module.
5. Multimedia module.
6. Project module.
7. Modular motion control module.

### 3.2 The iSanbot architecture

The architecture used for developing the iRobot application is a client-server one. At high level, there are two main components: a client, written in Java for Android, and a server, written in Java, and located on a dedicated machine. The client's goal is to interact with the user and to acquire data aimed at customizing the user-robot interaction, while the server represents the robot's "brain" in the cloud. More in detail, the architecture has been implemented as follow.

The client side, represented in the current scenario by Sanbot Elf hosting the iSanbot application, is compatible with any other robot equipped with a network connection and able of acquiring audio and video streaming. The server side, written in Java as well, acts as a glue between the various software components, each of which creates a different service (see Figure 2). In particular, the main services are: the emotion recognition, the face recognition, the user modeling and finally the speech-to-text service. The latter has been created using the Wit service[2] (a free Speech to Text service) and communication takes place through a simple HTTP request, sending the audio to be converted into text and waiting for the response. The face recognition, instead, has been developed in Python and implements the librarY "FaceRecognition.py"[3], while the emotion recognition service has been developed in C# and implements the libraries of Affectiva[4]. Given the diversity of services, a single programming language that would allow all libraries to be implemented was not found, so we created a central server written in Java, which has the task of delegating requests to the specific components. Furthermore, it has the role of saving data to a central database, after the various processing of each component. There are therefore two types of communication: the one between client and Java server, and other ones between the Java server and the various internal and external components.

The application starts with a natural language conversation that has the goal of knowing the user (e.g., face, name, sex, age, interests and predominant mood, etc.) and then recovering all the acquired and elaborated information from the user modeling component for the following interactions, so that the robot will be able to adapt its behavior, and eventually deliver recommendations and/or stimulate correct behavior with its persuasion [20].

### 3.3 Face recognition service

First, we implemented the face recognition service. Sanbot, as soon as the user approaches, welcomes and takes a picture of her, and sends it to the Java server, which delegates the Python component to the user's recognition. If the user has not been recognized, a new user will be inserted into the user database. The user is associated to a sets of attributes, such as name, surname, age, sex, profession, interests, moods, etc. to be filled in the following. Some of the attribute indeed may be associated with an explicit question, while other may be implicitly inferred. Thus the client, during the user registration phase, asks the server what the next question is and then it reads it to the user. Once the user has answered, the audio is sent to the Wit server, converted into text and, one returned, saved into the corresponding attribute associated with the question. For example Sanbot, if does not recognized the user, asks the user her name, then, before taking her a picture, it asks the permission to (*Do you want me to remember you the next time i see you*?), then it asks her profession, and her favorite color and sport, and stores all these information on the user DB.
The detected predominant emotions will be inferred and stored, as well as other biometric data,as described in the following.

---

[2] https://wit.ai/
[3] 3https://github.com/ageitgey/facerecognition, by Adam Geitgey
[4] https://www.affectiva.com/

On the other hand, if the face recognition component recognizes the user, all her stored information are recovered and the user is welcomed (e.g., *Hello Cristina! Do you feel better today?*) as well the next question to read. When there are no more questions, the conversation ends.

As far as the robot adaptive behavior is concerned, it is now based on simple rules, which are now used for demonstration purposes and which will be then modified and extended in our future work. For example, in the current demo, the robot performs a more informal conversation with young people and a more formal one with older people, it is more playful with young people and more serious with older people. As for emotions, if it sees a sad face he says a joke or plays a song, while he smiles and congratulates the smiling people.

For accessibility reasons and even better understanding, the application replicates all the conversation transcription on the Sanbot tablet screen, both the phrases spoken by Sanbot and those ones acquired by the user.

### 3.4 Emotion recognition service

During conversation with the user, the emotion recognition component is running. This component has been realized through a continuous streaming of frames that is directed towards the C# component, which analyzes each frame and extracts the predominant emotions. After that, the server sends a response via a JSON object to the client, which can therefore adapt its behavior and its conversation according to the rule associated to the dominant emotion. For instance, in our application, for demonstration purposes, Sanbot also changes its facial expression with the expression associated with user's mood.

Currently Sanbot is able to recognize six plus one emotions, also known as Ekman's [3] universal emotions and they are: sadness, anger, disgust, joy, fear, surprise and contempt.

The modeling of the user affective state is part of the user modeling component, and has been implemented by using a hashmap that has the name of the emotion as a key (e.g. Joy) and a data structure, containing the number of the occurrences and the intensity of the emotion (Affectiva returns a value for each detected emotion ranging from 0 to 99 depending on the perceived intensity). For example we could have Joy as hashmap containing 2 values, namely 2 and 104 (<Joy, 2, 104>): the first one represents the number of occurrences, while the second represents the sum of perceived intensity values. This data structure, called *totalEmotions*, is updated for each frame, so that, at the end of the conversation, we will have the story of all the emotions experienced and expressed in terms of sum of occurrences and sum of intensity. In this way it is then possible to associate the user to a predominant emotion and therefore to plan a behavior consistent with what has been detected.

The predominant emotion of a given interaction will be the one with a higher certainty factor associated. This certainty factor is obtained as the ratio between the the number of occurrences of the emotion weighted by its intensity and the total number of frames detected during the interaction.

On the basis of the predominant emotion, returned after the greetings, Sanbot will adapt its behavior accordingly.

The Affectiva SDK and API also provide estimation on gender, age range, ethnicity and a number of other metrics related to the analyzed user. In particular we use the information on gender and age range to implicitly acquire this data about the user and store them on the database. Since also this information is returned with a perceived intensity level, we use the same certainty factor calculation described above.

### 3.5 User modeling

Currently, we have created a very simple user model organized as a set of feature-value pairs, and stored in a database. The current stored features are: user name, age range, gender, favourite sport, favorite color, profession, number of interaction, predominant emotions. Some of the user acquired information are explicitly asked to the user (name, profession, sport, color), other ones (age, sex, emotions) are inferred trough the emotion recognition component described above. Please notice that explicit information as favorite color and sport are used for mere demonstration purposes. In the future we would like to extend the user features with other inferred features such as kind of personality (e.g. Extraversion, Agreeableness, Neuroticism, etc), kind of user dominant emotion, stereotypical user classification depending on socio-demographic information as sex, age and profession, and so on. Concerning this last point, we will base our stereotypes on shared social studies categorizing somehow the user's interests and preferences on the basis of socio-demographic information. We would also like to import part of the user profile by scanning social web and social media [17], with the user's consent.

## 4 Conclusion

Our cloud-based service to model the user and her emotion is just at the beginning. We are now working with the Pepper robot to replicate the same scenario implemented in Sanbot.

We are also linking the user database with official information stored in the Computer Science Dept. web site and regarding activities and details of the employees. Information as the number of office, office hours of the professor, class schedule, etc, can be sent to the user modeling component (through a REST service) in order to enrich the information available to the robot for its conversations with users.

In the future we would like to replace Affectiva with a similar service we are developing, extend the user model and integrate more sophisticated rules of inferences and adaptation, improve the component of natural language processing.